\documentclass{article}
\usepackage{spconf,amsmath,epsfig}
\usepackage{amsfonts}
\usepackage{amssymb}
\usepackage{xcolor}
\usepackage{url} 
\usepackage[font=small,labelfont=bf]{caption}
\usepackage{multirow}

\title{Plant detection from ultra high resolution remote sensing images: \\ A Semantic Segmentation approach based on fuzzy loss}

\name{S. Pande$^1$, B. Uzun$^1$, F. Guiotte$^{1,2}$, M.T. Pham$^1$, T. Corpetti$^3$, F. Delerue$^4$ and S. Lefèvre$^1$ \thanks{This research work is supported by ANR SixP project (ANR-19-CE02-0013). Corresponding author: shivam.pande@irisa.fr}}
\address{$^1$Université Bretagne Sud, UMR IRISA 6074, F-56000 Vannes, France \\
$^2$ Avion Jaune, Saint-Clément-de-Rivière, F-34980, France \\
    $^3$ LETG UMR 6554, CNRS, F-35000 Rennes, France \\
    $^4$ Univ. Bordeaux, CNRS, Bordeaux INP, EPOC, UMR 5805, F-33600 Pessac, France}

\begin{document}
%
\maketitle
\begin{abstract}
In this study, we tackle the challenge of identifying plant species from ultra high resolution (UHR) remote sensing images. Our approach involves introducing an RGB remote sensing dataset, characterized by millimeter-level spatial resolution, meticulously curated through several field expeditions across a mountainous region in France covering various landscapes. The task of plant species identification is framed as a semantic segmentation problem for its practical and efficient implementation across vast geographical areas. However, when dealing with segmentation masks, we confront instances where distinguishing boundaries between plant species and their background is challenging. We tackle this issue by introducing a fuzzy loss within the segmentation model. Instead of utilizing one-hot encoded ground truth (GT), our model incorporates Gaussian filter refined GT, introducing stochasticity during training. First experimental results obtained on both our UHR dataset and a public dataset are presented, showing the relevance of the proposed methodology, as well as the need for future improvement.
\end{abstract}
\section{Introduction}
\label{sec:intro}
\begin{figure}[htb]
\begin{minipage}[b]{1.0\linewidth}
  \centering
  \centerline{\includegraphics[width=8.35cm]{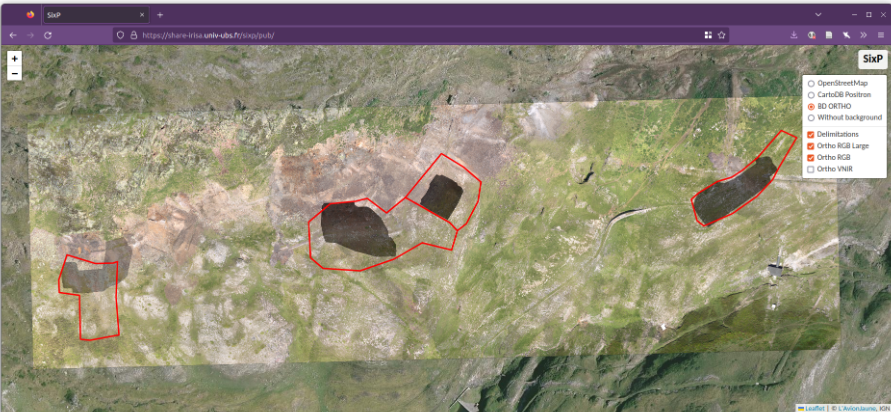}}
\end{minipage}
\caption{Raw RGB image over Chichoue site (one of the eight sites).}
\label{fig:data_img}
\end{figure}

Recent advancements in sensing technologies have significantly boosted research in the remote sensing community. With improved sensors, a vast amount of geospatial data from multiple sources and modalities is now available at ultra-high resolution (UHR). Land-cover mapping remains one of the most common yet challenging problems, and the challenges increase with UHR data due to high dimensionality, labeling costs, and large geographical areas \cite{horning2020land, sohl2024comparative}. In this study, we address a similar problem of plant species identification, which is a primary objective of our project: Positive Plant-Plant interactions and spatial Patterns in Pyrenean Post-mine tailings (SixP project). We collected ultra-high resolution multispectral (RGB and near-infrared, but only RGB is used here) imagery from a complex, heterogeneous study site in France for vegetation mapping and plant identification for the year 2020. We approach plant identification as a semantic segmentation problem due to its simpler implementation and high accuracy. Semantic segmentation has been widely used in remote sensing for various tasks such as swamp detection and crop delineation \cite{yuan2021review, lv2023deep}. However, most existing work benefits from precise ground truth with clearly defined boundaries, which is crucial for supervised training, especially with cross-entropy-based loss. Unlike these datasets, our dataset lacks clear delineation among different plant species due to natural overlapping, as evident in the subcentimetric image resolution.

In our dataset, reference data is obtained from field surveys, showing imperfect alignment with the images. Notable disparities exist between the plant representations in the images and the annotations, exacerbated by the simplification of plant shapes into discs. Further inconsistencies arise from differences between the drone-acquired, orthorectified images and the manually surveyed plant data, including species identification, GPS positions, and measured diameters. Temporal discrepancies also occur due to the interval between drone flights and species surveys, and potential labeling errors during manual data entry. Errors are introduced by systematic species identification practices, where plant manipulation and the discovery of previously unnoticed individuals during nadir visual observations add to inaccuracies. Technical limitations in creating mosaics also introduce errors, particularly in orthorectifying visible images using digital terrain models (DTM) with potential elevation discrepancies, affecting pixel projection positions from original photos onto orthorectified images. Due to errors in ground truth labels and the overlap among different plant species, semantic segmentation becomes highly challenging. To address these challenges, we introduce stochasticity in the ground truth during model training. We convolve our ground truth maps with a Gaussian kernel of predetermined mean and standard deviation, softening the labels and allowing the model to capture data uncertainty. In this setup, we treat the segmentation problem as a regression problem, using distribution rather than one-hot encoded labels. We introduce a new loss function, called \textit{fuzzy loss}, that can be formulated either using a distance metric (like mean squared error) or a similarity metric (like cross-entropy) between the predicted probabilities and the Gaussian-convolved ground truths.
To our knowledge, such an approach has been adopted for density estimation and object counting \cite{wan2020kernel,singh2023object}, but not in the context of semantic segmentation in the literature.

\begin{figure}[htb]
\begin{minipage}[b]{0.95\linewidth}
  \centering
  \centerline{\includegraphics[width=6.5cm]{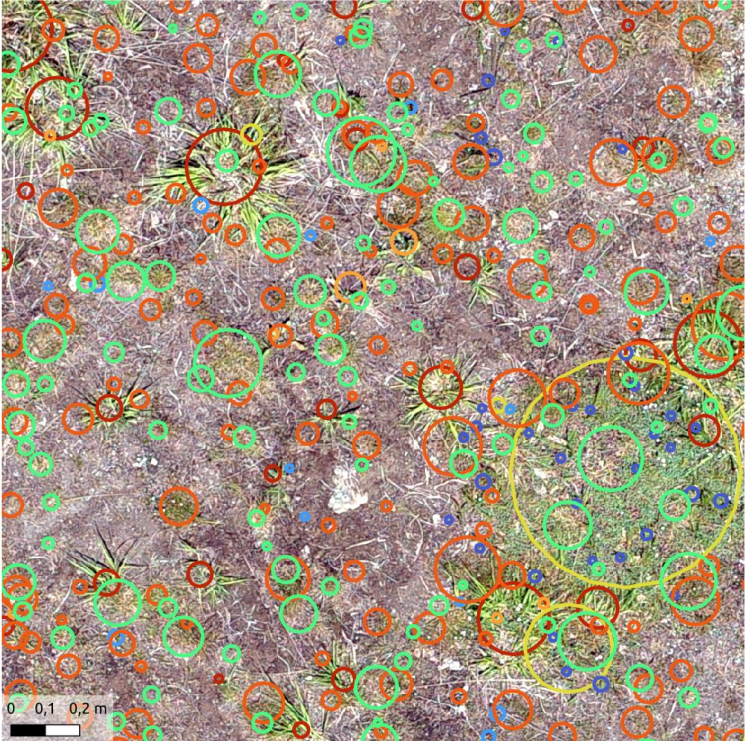}}
\end{minipage}
\caption{SixP dataset with RGB colour composite with overlapping circular ground truth for plants.}
\label{fig:sixpdata}
\end{figure}

The contributions of this research are enlisted below:
\begin{itemize}
  \item We introduce a novel dataset with ultra-high resolution RGB images from remote sensing domain for plant species identification, acquired in a realistic scenario with field observations.
  \item We approach plant identification as a semantic segmentation problem for efficiency and accuracy reasons. 
  \item To tackle the problem of overlapping classes in the segmentation masks, we propose a novel fuzzy loss, that brings in the notion of stochasticity in the GT labels by convolving them with a Gaussian kernel. 
\end{itemize}
\begin{figure*}[t!]
  \centering
  \centerline{\includegraphics[width=17cm]{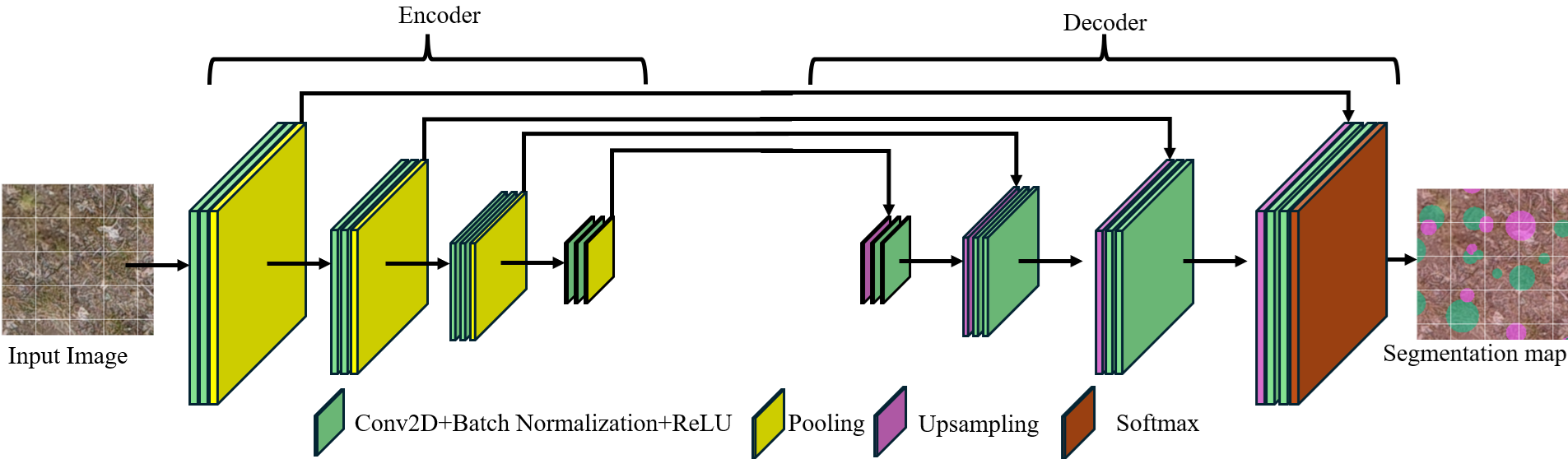}}
  \caption{Schematic of the U-Net based architecture for semantic segmentation.}\medskip
  \vspace{-0.5cm}
  \label{fig:Diagram}
\end{figure*}

\section{Methodology}
\label{sec:method}
We discuss here the datasets used in our study, the data preprocessing aspects, and the creation of fuzzy labels leading to the implementation of the semantic segmentation model. 

\subsection{Dataset}


\textbf{Our SixP dataset}: The data, acquired from UHR drone imagery and field surveys, includes visible (red, green, blue, \emph{i.e.,} RGB) and multispectral (RGB + near-infrared) photographs, but we only employ RGB data in this study. These images are orthorectified, mosaicked, and georeferenced, achieving 2-3 mm per pixel resolution. Data collection spanned eight zones in a complex study site in the Pyrenees, France. Field surveys provided reference data, with ecologists recording plant positions, identifications, and characteristics within quadrats of 1 to 25 m$^2$. Multiple quadrats were sampled per site. Raw plant data include differential GPS (DGPS) positions, species information, diameters, and area identifiers, with some plants defined by polygons. Fig. \ref{fig:data_img} shows the raw RGB dataset, and Fig. \ref{fig:sixpdata} presents the ground truth with ring-shaped bounding boxes.

\textbf{The Weed Dataset}: Since our SixP dataset is not yet publicly available, we rely on a similar public dataset for weed detection \cite{jai_dalmotra_2023}. This collection features diverse weed species from various environments, climates, and conditions, reflecting real-world detection challenges. Each high-resolution image is meticulously annotated to identify weed presence and location, aiding in training and assessing computer vision models. The dataset also includes metadata such as location, date, and plant information to provide additional contexts \cite{jai_dalmotra_2023}.

\subsubsection{Data preprocessing}

For both datasets, the ground truths are provided in the form of circular or elliptical box annotations around the plant species. Hence, to formulate the segmentation problem, the rings are converted to the segmentation masks. In our case, we treat the plant identification problem as a binary segmentation task, where the plant species represent the foreground, while the rest of the image is considered as the background. We tackle the semantic segmentation task from two different perspectives. The first one is the conventional semantic segmentation problem, where the values inside each ring are homogeneously declared as the foreground. In the second approach, we use a Gaussian mask and convolve it over the image such that the values closer to the centre of the plant have a higher magnitude in the GT, while as we move away from the plant centre and towards the periphery, the certainty of a the pixel being denoted as plant decreases. This is illustrated Fig. \ref{fig:sixp_segmap} (d) and Fig. \ref{fig:w_segmap} (d). After creation of the GT, the images were divided into smaller patches of 640 $\times$ 640 and 320 $\times$ 320 pixels, for SixP and Weed dataset, respectively.

\subsection{Problem definition}
After pre-processing the dataset, let us consider a set of UHR RGB images denoted as $\mathcal{X} = \mathbf{x}_{i=1}^n$ such that $\mathcal{X} \in \mathbb{R}^{P \times Q \times B}$. Here, $n$ is the number of samples, while $P$ and $Q$ are spatial dimensions, and $B$ is the number of channels. The corresponding GT for the images is given as $\mathcal{Y} = \mathbf{y}_{i=1}^n$ such that $\mathcal{Y} \in \mathbb{R}^{P \times Q \times C}$. Here, $C$ represents the total number of classes. The entire problem is posed as a semantic segmentation task such that each pixel in $\mathcal{X}$ can be mapped to a corresponding class in $\mathcal{Y}$.   

\subsection{Creation of fuzzy labels}
To account for the spatial imprecision inherent in class delineation, we present a novel approach involving the modeling of spatial confidence within the reference data. Our method entails convolving the pixel membership to a class with a Gaussian kernel, leveraging its ability to represent the spatial probability of class membership based on the standard deviation of DGPS errors. The kernel is represented in Eq.~\eqref{eq:GC}, where $\mathbf{y}_p$ and $\mathbf{y}_q$ represent the spatial locations in the image, while $\sigma_y$ is the standard deviation that is treated as a hyperparameter: 
\begin{equation}
G_{2D}(\mathbf{y}_p,\mathbf{y}_q,\sigma_y)=\frac{1}{2\pi{\sigma}^2}\exp^{-\big(\frac{\mathbf{y}_p^2+\mathbf{y}_q^2}{\sigma_y^2}\big)}
\label{eq:GC}
\end{equation} 
The modified GT can be represented as ${\mathbf{y}}_G = G_{2D}(\mathbf{y})$.
The proposed implementation augments the cost functions with an initial step aiming at ensuring the integration of spatial confidence modeling. The efficiency of convolutions, particularly on GPUs, is harnessed for this purpose. Additionally, the 2D decomposition of Gaussian kernels proves advantageous, leading to significant acceleration in computational speed. 

\subsection{Model architecture}
In this study, we exclusively used the U-Net network \cite{ronneberger2015u} (shown in Fig. \ref{fig:Diagram}). This choice was motivated by its versatility and significant presence in the state-of-the-art, but let us emphasize that our contributions can be implemented with other models as well. The network consists of an encoder $\mathcal{E}$ to downsample the original image to a bottleneck representation $\mathcal{E}(\mathbf{x}^i)$, and a decoder $\mathcal{D}$ to upsample the bottleneck representation to $\mathcal{D}(\mathcal{E}(\mathbf{x}^i))$. The last layer of the decoder represents a softmax layer, that outputs the probabilities for the different pixels in the images to belong to the different classes. Since we are working with fuzzy labels, we present a modified fuzzy loss function that converts the classification aspect of the segmentation task to a regression based setting. For the training, we use different loss functions, such as binary cross-entropy (BCE), mean-squared error (MSE) and cosine similarity (CS) between the fuzzy GT and the calculated probabilities from the network (see Eq. \ref{eq:bce}, \ref{eq:mse} and \ref{eq:cs} for the respective losses), on which the model is trained. 
\begin{equation}
loss_{CE} = -\mathbf{y}^i_G \log{\mathcal{D}(\mathcal{E}(\mathbf{x}^i)})
\label{eq:bce}
\end{equation} 
\begin{equation}
loss_{MSE} = (\mathbf{y}^i_G - {\mathcal{D}(\mathcal{E}(\mathbf{x}^i)}))^2
\label{eq:mse}
\end{equation} 
\begin{equation}
loss_{CS} = 1 - \frac{\mathbf{y}^i_G \cdot \mathcal{D}(\mathcal{E}(\mathbf{x}^i))}{\|\mathbf{y}^i_G\|\|\mathcal{D}(\mathcal{E}(\mathbf{x}^i))||}
\label{eq:cs}
\end{equation} 

In the inference phase, the validation/test images are sent to the trained model and the corresponding class probabilities are obtained, from which is performed class assignment. 

\section{Experiments and Preliminary Results}
\label{sec:EnR}
In this section, we will discuss the experimental setup and the preliminary results of our investigation.

\subsection{Training protocols and evaluation} 
The optimization is carried out using Adam optimizer with Nesterov momentum \cite{dozat2016incorporating} with an initial learning rate of 0.0005 and a gradual learning rate decay. All the models are trained on a Nvidia A6000 GPU. For evaluation, we use overall accuracy, classwise accuracy, Cohen's kappa and F1-score, for the conventional segmentation case. In case of fuzzy loss based segmentation, we use regression based metrics such as mean squared error and cosine similarity. 
\begin{table}[ht]
\centering{\scriptsize
\caption{\label{tab:cls_perf} Quantitative evaluation of segmentation performance in conventional setting (all values in \%). OA stands for Overall Accuracy.}
\begin{tabular}{c|c|c|c|c|c}
     & Background & Plants & OA & Kappa $\kappa$ &F1 score\\
     \hline
SixP &$94.31$&$48.75$&$91.49$&$37.07$&$41.55$\\
\hline
Weed &$99.30$&$48.57$&$95.95$&$59.33$&$61.31$\\
\end{tabular}}
\end{table}
\subsection{Results and discussion} 

\begin{figure}[htb]
\begin{minipage}[b]{1.0\linewidth}
  \centering
  \centerline{\includegraphics[width=\textwidth]{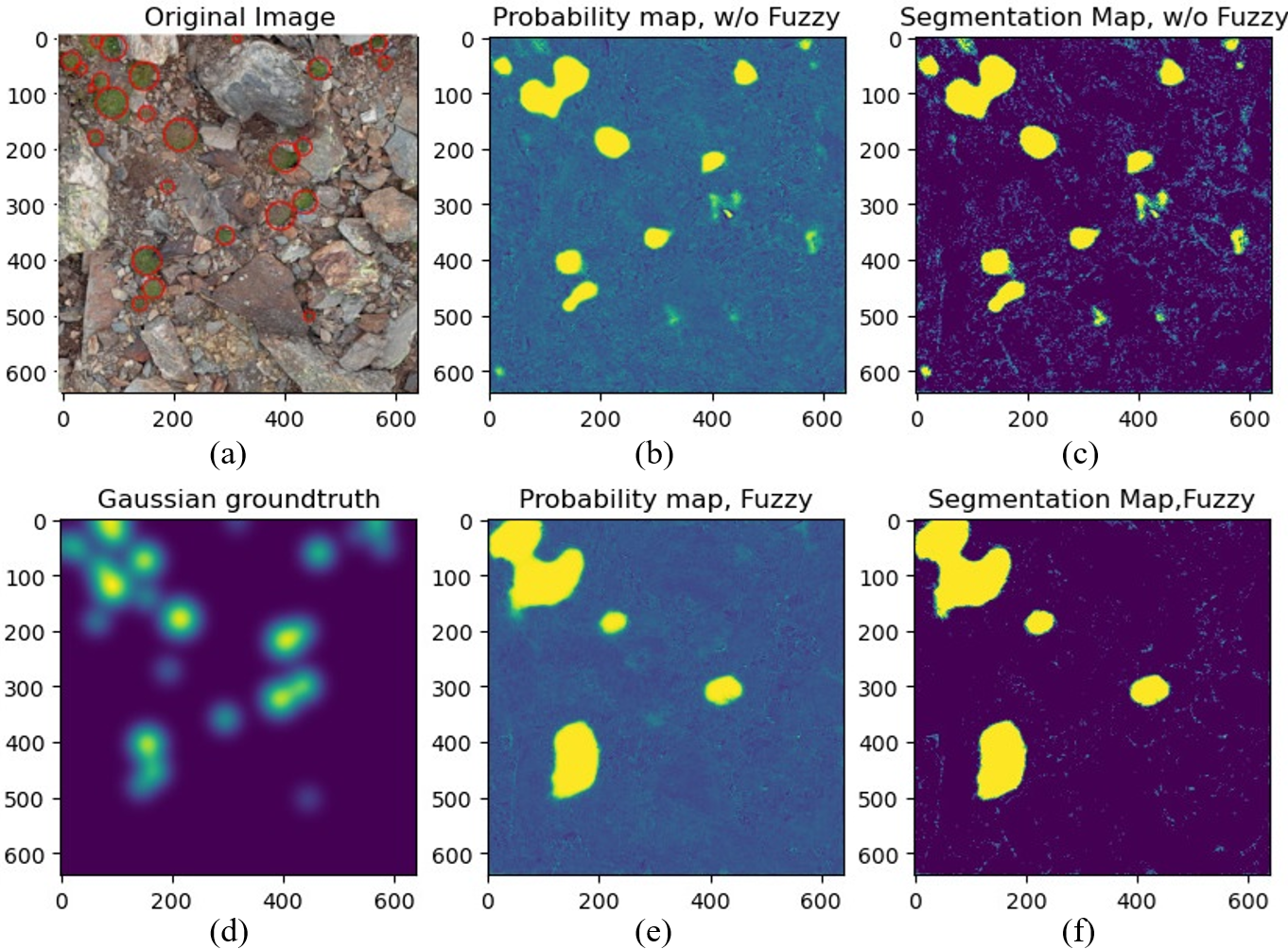}}
\end{minipage}
\caption{Visual illustration on the SixP dataset: original image (a), probability map (b) and segmentation map (c) in the classical case, Gaussian-convolved groundturh (d), and probability map (e) and segmentation map (f) with our fuzzy loss.}
\label{fig:sixp_segmap}
\end{figure}
Table \ref{tab:cls_perf} illustrates the semantic segmentation outcomes in the conventional context for both datasets. The overall accuracy stands at 91.49\% for the SixP dataset and 95.95\% for the Weed dataset. However, despite these high accuracies, the $\kappa$ and F1-scores are relatively diminished due to significant class imbalances. This imbalance is evident in the classwise accuracy of the foreground (plants) for both datasets. Table \ref{tab:reg_perf} displays results employing a fuzzy loss-based approach, where the cosine loss yields optimal performance for the SixP dataset, while the cross-entropy loss proves most effective for the Weed dataset.
\begin{figure}[htb]
\begin{minipage}[b]{1.0\linewidth}
  \centering
  \centerline{\includegraphics[width=\textwidth]{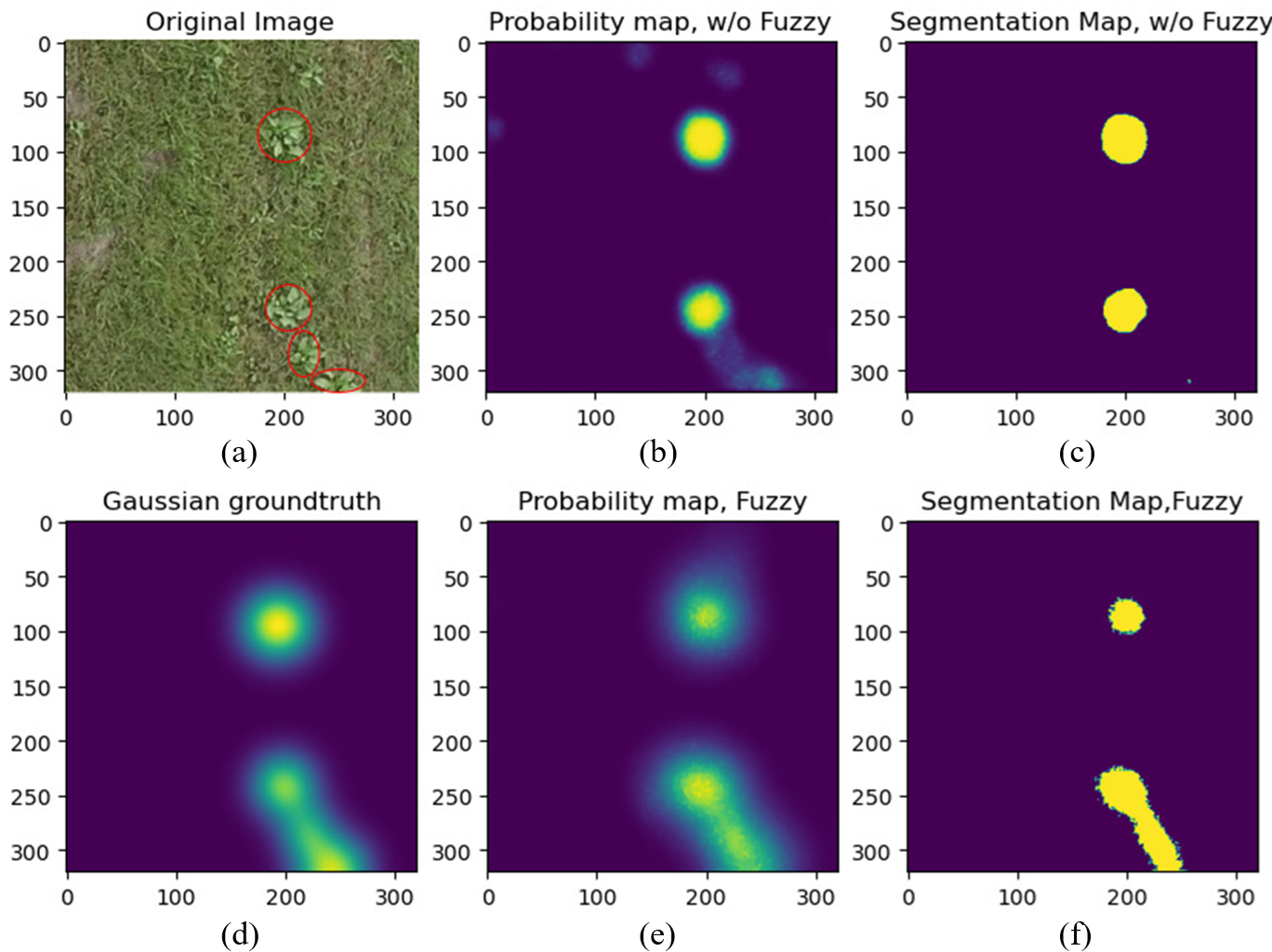}}
\end{minipage}
\caption{Visual illustration on the Weed dataset: original image (a), probability map (b) and segmentation map (c) in the classical case, Gaussian-convolved groundturh (d), and probability map (e) and segmentation map (f) with our fuzzy loss.}
\label{fig:w_segmap}
\end{figure}

Figs. \ref{fig:sixp_segmap} and \ref{fig:w_segmap} exhibit the visual outcomes of segmentation methods applied to the SixP and Weed datasets, respectively. These figures showcase segmentation and probability maps derived from logits, emphasizing the accuracy of predictions at the centers of plant species. Particularly in the Weed dataset (Fig. \ref{fig:w_segmap}), the impact of the fuzzy label-based approach is evident, with the probability density map effectively discerning weeds, unlike the traditional model. This can be noticed by comparing the bottom part of Fig.~\ref{fig:w_segmap} (b) and (e). However, such distinctions are less pronounced in the SixP dataset, possibly due to highly imbalanced classes and increased label noise, which pose challenges in segmentation. Notably, smaller plants are entirely overlooked, suggesting that Gaussian smoothing might suppress crucial details alongside label noise.

\begin{table}[!ht]
    \centering{\scriptsize
    \caption{\label{tab:reg_perf} Quantitative evaluation of segmentation performance in the fuzzy setting. Conversely to the standard setting using classification metrics, regression metrics are employed here (i.e., the lower the better).}
    \begin{tabular}{l|l|l|l|l}
         & Metrics& MSE Loss& Cosine Loss& Cross Entropy\\ \hline
        \multirow{2}{*}{SixP} & MSE & $0.0784$ & $0.0772$ & $0.1326$ \\ 
         & Cosine Sim.& $0.9161$ & $0.9181$ & $0.8485$ \\ \hline
        \multirow{2}{*}{Weed} & MSE & $0.0196$ & $0.0138$ & $0.0092$ \\ 
         & Cosine Sim.& $0.9789$ & $0.9861$ & $0.9896$ \\ 
    \end{tabular}}
\end{table}

\section{Conclusion}
\label{sec:conclusion}
In this research, we have introduced a new RGB dataset (a.k.a. the SixP dataset) on UHR remote sensing images for the task of plant species identification on a large scale. We have formulated the plant species detection as a semantic segmentation problem for efficient and accurate identification. Simultaneously, to tackle the challenges of noisy and overlapping GT labels, we have introduced a fuzzy loss function, that morphs the hard-coded GT to a more stochastic representation using Gaussian kernel. We have showcased the performance of the models on  two plant species detection datasets (our SixP dataset and the public Weed Image Detection dataset), while the strengths and weaknesses of both the approaches are identified. We conclude from these first experiments that, due to a high class imbalance and the severity of noisy labels in the SixP data, further research is required to ensure better identification of plant species in such a complex but realistic ultra-high resolution imaging scenario.



\newpage
\bibliographystyle{IEEEbib}
{\footnotesize\bibliography{refs}}

\end{document}